\documentclass[10pt]{article}
\pdfoutput=1
\usepackage[margin=1in]{geometry}
\usepackage{graphicx}
\usepackage{booktabs}
\usepackage{amsmath,amssymb}
\usepackage{hyperref}
\usepackage{tabularx}
\usepackage{placeins}
\usepackage{microtype}
\usepackage{authblk}
\hypersetup{colorlinks=true,linkcolor=black,citecolor=black,urlcolor=blue}

\title{City-Conditioned Memory for Multi-City Traffic and Mobility Forecasting}
\author[1]{Du Wenzhang}
\affil[1]{Department of Computer Engineering, Mahanakorn University of Technology, International College (MUTIC), Bangkok, Thailand\\\texttt{dqswordman@gmail.com}}
\date{\today}

\begin{document}
\maketitle

\begin{abstract}
Deploying spatio-temporal forecasting models across many cities is challenging: traffic networks differ in size and topology, data volume per city can vary by orders of magnitude, and new cities often arrive with only a few weeks of logs. Existing deep traffic models are typically trained per city and backbone, making it costly to maintain separate predictors for each deployment. In this work we ask: can we design a single, backbone-agnostic layer that conditions on ``which city this sequence comes from'', improves accuracy in both full- and low-data regimes, and yields better cross-city adaptation, while requiring minimal changes to existing models?

We propose \textbf{CityCond}, a light-weight city-conditioned memory layer that can be attached to a broad family of spatio-temporal backbones. CityCond combines a city-ID encoder with an optional shared memory bank (CityMem). Given a city index and backbone hidden states, it produces city-conditioned features that modulate the backbone through residual connections. We instantiate this design with five representative backbones---GRU, temporal convolution networks (TCN), Transformers, graph neural networks (GNN), and spatial-temporal GCNs (STGCN)---and evaluate across three realistic regimes: (i) full-data training on METR-LA and PEMS-BAY, (ii) low-data training with only 5--50

Across 14+ model variants and three random seeds, CityCond yields consistent gains on traffic forecasting, with the largest improvements for high-capacity backbones. On full-data traffic, CityMem-Transformer reduces error by roughly one third compared to a shared Transformer baseline, and CityMem-STGCN improves over STGCN by 30--35
\end{abstract}


\section{Introduction}
Urban traffic and mobility forecasting are core building blocks of AI-powered smart cities. Accurate predictions of road speed, volume, or pedestrian trajectories enable congestion mitigation, dynamic routing, and safer planning for shared spaces. Over the past decade, deep spatio-temporal models---diffusion convolutional RNNs (DCRNN)~\cite{li2018dcrnn}, STGCN~\cite{yu2018stgcn}, Graph WaveNet~\cite{wu2019graphwavenet}, MTGNN~\cite{wu2020connecting}, adaptive graph RNNs~\cite{bai2020agrn}, among many others---have significantly advanced single-city traffic forecasting. However, large operators rarely serve a single urban system: they must deploy models across dozens of cities, each with different topology, sensor layouts, and data volumes.

A straightforward approach is to train a separate model per city and per backbone. This siloed paradigm does not scale. It duplicates training and maintenance cost, and it fails when a new city has only a short history of logs. Meanwhile, an emerging line of work explores cross-city transfer learning and meta-learning: MetaST learns a pattern-based spatio-temporal memory from multiple cities~\cite{yao2019metast}; RegionTrans (Wang et al.) aligns similar regions across cities for transfer~\cite{yao2019regiontrans}; ST-MetaNet (Pan et al.) uses deep meta learning for traffic prediction~\cite{panagopoulos2021stmetanet}; and Selective Cross-City Transfer (CrossTReS) re-weights source regions for transfer~\cite{jin2022selective}. These methods demonstrate that city-level knowledge can be shared, but they often require complex training loops or tightly coupled architectures that are hard to drop into existing systems.

In this work we explore a complementary question: \emph{Can we build a single, backbone-agnostic layer that conditions on city identity, can be plugged into existing spatio-temporal models with minimal code changes, and consistently improves performance across full-data, low-data, and cross-city regimes?}

We answer this by proposing \textbf{CityCond}, a city-conditioned memory layer. CityCond takes as input a city index, backbone hidden states, and (optionally) raw node features; it outputs a city-conditioned feature vector that is fused back into the backbone. Concretely, CityCond has two components: (i) a CityID encoder that maps a discrete city index to a learnable embedding; and (ii) CityMem, an optional shared memory bank with $K$ slots, queried by the city embedding and current hidden states via attention to produce a city-specific context vector. Because CityCond only depends on a city id and the backbone's feature tensors, it can be attached to a wide range of models without modifying their recurrence, convolution, or attention patterns.

We instantiate CityCond on three representative traffic backbones---GRU, TCN, Transformer---and two graph-based backbones---GNN-based speed prediction and STGCN~\cite{yu2018stgcn}. For each backbone we study three variants: \emph{Base} (a multi-city shared model), \emph{CityID} (Base plus a city embedding), and \emph{CityMem} (CityID plus the memory bank). Our experiments cover: (i) full-data training on METR-LA and PEMS-BAY; (ii) low-data training where each city provides only 5--50

\begin{figure}[htbp]
    \centering
    \includegraphics[width=0.8\linewidth,keepaspectratio]{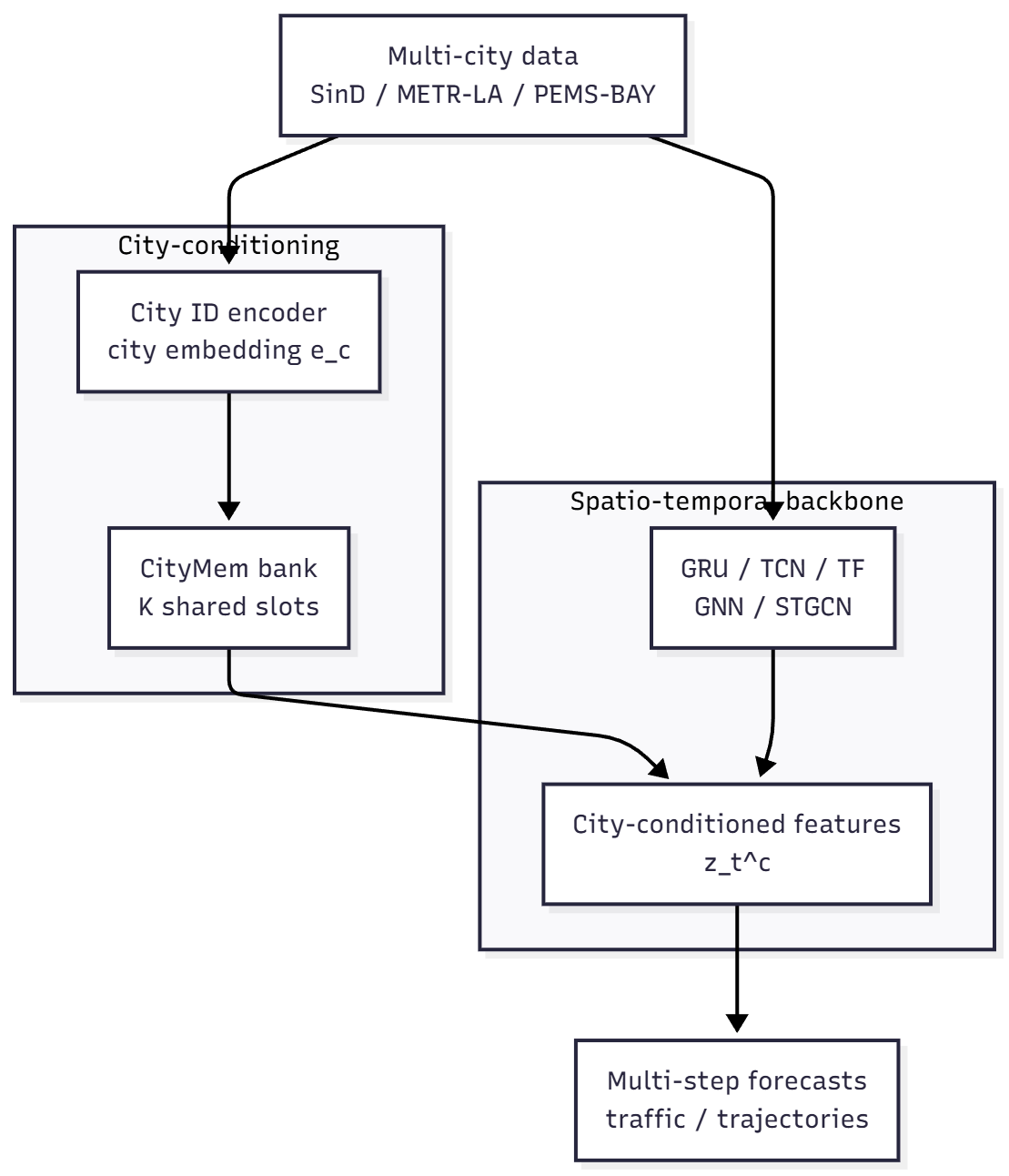}
    \caption{Overview of CityCond. A city index is embedded and optionally used to query a shared memory bank (CityMem). The resulting city-conditioned context is fused back into backbone hidden states through gated residual connections, enabling backbone-agnostic integration.}
    \label{fig:overview}
\end{figure}

Figure~\ref{fig:overview} gives an overview of the architecture. Figure~\ref{fig:traffic} and Figure~\ref{fig:transfer} summarize the main empirical trends. Tables~\ref{tab:datasets}--\ref{tab:tf_xcity} report detailed numbers.

\paragraph{Contributions.} Our main contributions are:
\begin{itemize}
    \item \textbf{City-conditioned memory as a design pattern.} We formalize CityCond as a general-purpose layer that augments existing spatio-temporal backbones with city-aware conditioning. CityCond can be implemented as a small module with a clear input--output interface, without altering backbone internals.
    \item \textbf{A unified multi-backbone, multi-regime benchmark.} We provide a systematic evaluation of city conditioning across four families of backbones (sequence-only RNN/TCN/Transformer, graph-based GNN/STGCN), three data regimes (full-data, low-data, cross-city few-shot), and three datasets (METR-LA, PEMS-BAY, SIND). All experiments use common training protocols and multi-seed evaluation.
    \item \textbf{Empirical characterization of when city conditioning helps.} We show that CityID and CityMem bring the largest gains for high-capacity backbones (Transformer, STGCN) in low-data and cross-city regimes, while simple LSTM backbones on human trajectories see more modest improvements. Our analysis highlights the interaction between backbone capacity, data budget, and city-conditioned memory.
\end{itemize}

\section{Problem Setup}
We consider a set of cities (or scenes) $C = \{c_1, \ldots, c_{|C|}\}$. For each city $c$ we observe multivariate time series indexed by discrete time steps $t = 1, \ldots, T_c$.

\subsection{Multi-city traffic forecasting}
For road traffic, each city $c$ has a fixed set of sensors (nodes) $V_c$ with $|V_c| = N_c$. At time $t$ we observe node-level features $x_{t,i}^c \in \mathbb{R}^{d_x}$ (e.g., speed) for node $i \in V_c$. The graph structure may be given by an adjacency matrix $A_c$ (GNN/STGCN) or implicitly handled in fully-connected backbones (Transformer).

Given a history window of length $L_h$, we write the past observations as
\[
    X_{t-L_h+1:t}^c = \{x_{\tau,i}^c\}_{\tau = t-L_h+1}^{t}, \quad i \in V_c.
\]
Our goal is to predict the next $L_f$ steps:
\[
    \hat{X}_{t+1:t+L_f}^c = f_{\theta}\bigl(X_{t-L_h+1:t}^c, c\bigr),
\]
where $f_{\theta}$ is a spatio-temporal forecasting model that may optionally depend on a city index $c$ through CityCond. We train $f_{\theta}$ to minimize mean squared error (MSE) or mean absolute error (MAE) over all cities and time steps.

\subsection{Multi-agent trajectory forecasting on SIND}
For intersection-level multi-agent trajectories, we use the SIND drone dataset~\cite{wei2024sind}, collected at a signalized intersection in Tianjin. It contains multi-agent tracks with 2D positions $p_{t,j}^c \in \mathbb{R}^2$ for agent $j$, covering several traffic participant types. We subsample fixed-length windows of length $L_h + L_f$. Given history positions $p_{t-L_h+1:t}^c$, we predict future positions $p_{t+1:t+L_f}^c$ for each agent using an LSTM-based encoder--decoder.

Unlike traffic, SIND does not involve a static road graph; agents move in continuous 2D space with complex interactions. We therefore treat SIND primarily as a sanity check for city conditioning in a different mobility domain, rather than as the primary target of CityMem.

\subsection{Backbone variants and regimes}
For each backbone we consider three variants:
\begin{itemize}
    \item \textbf{Base}: a single model trained jointly on all cities, without explicit city conditioning.
    \item \textbf{CityID}: Base plus a city embedding; the embedding is concatenated to input features or hidden states.
    \item \textbf{CityMem}: CityID plus a shared memory bank; CityMem is the full CityCond layer.
\end{itemize}
We evaluate in three regimes:
\begin{itemize}
    \item \textbf{Full-data}: use all available training data from each city.
    \item \textbf{Low-data}: uniformly subsample a fraction $\mathrm{frac} \in \{0.05, 0.10, 0.20, 0.50\}$ of training windows per city.
    \item \textbf{Cross-city few-shot}: pre-train a model on a source city, then fine-tune on a target city using a short adaptation window, evaluating both pre- and post-adaptation performance.
\end{itemize}

\section{City-Conditioned Memory Layer}
We now describe CityCond, focusing on its interface and how it integrates with different backbones.

\subsection{CityID conditioning}
We first introduce a simple city-ID conditioning baseline. Let $c \in C$ be a city index and $e_c \in \mathbb{R}^{d_c}$ a learnable embedding vector. For each time step $t$ and node $i$, we form augmented inputs by concatenating $e_c$:
\[
    \tilde{x}_{t,i}^c = [x_{t,i}^c; e_c].
\]
For sequence-only backbones (GRU/TCN/Transformer), we concatenate $e_c$ to each time step's feature vector. For graph-based backbones, we either concatenate $e_c$ to node features or to pooled graph features, depending on the architecture. This CityID variant is parameter-efficient (only $|C| \times d_c$ extra parameters) and serves as a strong baseline for ``hard'' city conditioning without memory.

\subsection{CityMem: city-conditioned memory bank}
CityMem extends CityID with a shared memory bank that captures reusable patterns across cities.

\paragraph{Memory structure.} CityMem maintains a learnable parameter matrix $M \in \mathbb{R}^{K \times d_m}$, where each row $m_k$ is a memory slot. Memory slots are shared across all cities and backbones using CityCond.

\paragraph{Query construction.} Given backbone hidden states $h_{t,i}^c$ for node $i$ at time $t$, we first compute a city-pooled hidden state $h_t^c$. For sequence-only models, we use mean pooling over nodes (or directly use the sequence hidden state if there is no node dimension). For graph-based models, we apply mean or max pooling over nodes in the graph. We then form a query vector by combining city embedding and pooled hidden state:
\[
    q_t^c = \phi_q\bigl([e_c; h_t^c]\bigr),
\]
where $\phi_q$ is a small MLP.

\paragraph{Attention readout.} We compute attention weights over memory slots:
\[
    \alpha_{t,k}^c = \frac{\exp(\langle q_t^c, m_k \rangle)}{\sum_{k'=1}^{K} \exp(\langle q_t^c, m_{k'} \rangle)}.
\]
The memory readout is
\[
    m_t^c = \sum_{k=1}^{K} \alpha_{t,k}^c \, m_k.
\]

\paragraph{Fusion with backbone states.} We fuse memory readout into backbone states via a gated residual:
\[
    g_{t,i}^c = \sigma\!\left(W_g [h_{t,i}^c; m_t^c]\right), \qquad
    \tilde{h}_{t,i}^c = h_{t,i}^c + g_{t,i}^c \odot W_m m_t^c,
\]
where $\sigma$ is a sigmoid, $W_g$ and $W_m$ are learnable matrices, and $\odot$ is elementwise multiplication. The augmented states $\tilde{h}_{t,i}^c$ are then passed to subsequent backbone layers. If we remove the dependence on $e_c$ in $q_t^c$, CityMem reduces to a global pattern memory shared across all cities; we treat this as an ablation in the appendix.

\subsection{Backbone-agnostic integration}
A key design goal is that CityCond should be backbone-agnostic: it should require minimal changes to existing models. We use a simple contract. Inputs to CityCond include the city index $c$, node-level features or hidden states $\{h_{t,i}^c\}$, and optional raw inputs $\{x_{t,i}^c\}$. Outputs are augmented features $\tilde{h}_{t,i}^c$ with the same shape as $h_{t,i}^c$. This enables us to wrap existing backbone layers without modifying their internal recurrence or convolution.

For Transformer-based traffic models, we consider an encoder-only architecture operating on flattened node--time tokens. CityID concatenates $e_c$ to each token's input. CityMem is inserted after the first encoder block: we pool hidden states to form $h_t^c$, compute $m_t^c$, and fuse it back into all tokens at step $t$ using the gated residual above. For sequence-only GRU and TCN backbones, node features are aggregated into a single per-time-step vector. CityID is concatenated to each time-step input. CityMem again operates after an early block, pooling across time or channels to form $h_t^c$ and fusing $m_t^c$ into time-step states. For graph-based traffic backbones (our GNN and STGCN implementations), hidden states $h_{t,i}^c$ live on nodes. We pool across nodes to obtain $h_t^c$ for querying memory, then broadcast the result back to nodes and apply gated fusion. This preserves the original spatial message-passing structure~\cite{yu2018stgcn,wu2019graphwavenet,wu2021survey}.

\subsection{Complexity}
CityCond adds $\mathcal{O}(|C| d_c + K d_m)$ parameters and a small number of linear layers. In our experiments, this increases parameter count by less than 5\% and training time by under 10\% for all backbones. Compared to meta-learning schemes that require additional inner-loop optimization~\cite{yao2019metast,panagopoulos2021stmetanet,jin2022selective}, CityCond retains the simplicity of standard supervised training.
\section{Experimental Setup}
\subsection{Datasets}
METR-LA and PEMS-BAY are standard traffic benchmarks consisting of sensor networks in Los Angeles and the Bay Area, respectively, with 5-minute aggregated speed readings~\cite{li2018dcrnn,yu2018stgcn,wu2019graphwavenet}. We follow common practice and use 12-step histories to predict 12-step (METR-LA) or 6-step (PEMS-BAY) horizons, applying per-node $z$-score normalization. SIND is a drone dataset from a signalized intersection in Tianjin~\cite{wei2024sind}, providing tracked 2D positions for multiple road-user types at roughly 10~Hz. We subsample to a constant frame rate and construct windows of 20 observed and 10 future steps. Table~\ref{tab:datasets} summarizes dataset statistics, history/horizon lengths, and the number of cities/scenes.

\begin{table}[htbp]
    \centering
    \small
    \caption{Datasets and forecasting setups. History length $L_h$ and horizon $L_f$ follow Section~\ref{sec:setup}. Traffic datasets use 5-minute aggregated speeds on fixed road graphs; SIND uses multi-agent trajectories from a drone view without a static graph.}
    \label{tab:datasets}
    \begin{tabularx}{\linewidth}{l l X X c c}
        \toprule
        Dataset & Domain & Cities/Scenes & Sampling & $L_h$ & $L_f$ \\
        \midrule
        METR-LA  & Traffic speed & Los Angeles (1 city) & 5-minute aggregation & 12 & 12 \\
        PEMS-BAY & Traffic speed & Bay Area (1 city) & 5-minute aggregation & 12 & 6 \\
        SIND     & Multi-agent trajectories (intersection) & 1 signalized intersection (Tianjin) & $\approx$10~Hz (drone, subsampled) & 20 & 10 \\
        \bottomrule
    \end{tabularx}
\end{table}

\subsection{Backbones and CityCond variants}
We instantiate five backbones: GRU (a two-layer gated recurrent unit network over flattened sensor features), TCN (a temporal convolutional network with dilated convolutions), Transformer (a 4-layer encoder with multi-head self-attention), GNN (a graph neural network with message passing over the sensor graph), and STGCN (a spatial-temporal graph convolutional network similar to~\cite{yu2018stgcn}). For each backbone we train three variants: Base, CityID, CityMem. For SIND, we use an LSTM encoder--decoder with Base and CityID variants.

\subsection{Data regimes and training protocol}\label{sec:setup}
For each city and backbone, we consider: (i) full-data using all training windows; (ii) low-data by randomly sampling 5\%, 10\%, 20\%, or 50\% of training windows while holding out a fixed validation set; and (iii) cross-city transfer, where Transformer variants are pre-trained on one city (e.g., METR-LA) then fine-tuned for a fixed number of steps on the other (PEMS-BAY) using a small labeled set, evaluating both pre- and post-adaptation performance. We train all models with Adam, using early stopping on validation error. We run three random seeds (13, 21, 42) per configuration and report mean and standard deviation.

\subsection{Evaluation metrics}
For traffic we report MSE and MAE averaged over all nodes and forecast horizons. For SIND we use final-displacement error (FDE) and average-displacement error (ADE) in meters, following standard practice~\cite{salzmann2020trajectronpp,gupta2018socialgan,mohamed2020socialstgcnn}.
\section{Results}
\subsection{Full-data traffic forecasting}
Table~\ref{tab:traffic_full} summarizes full-data results on METR-LA+PEMS-BAY. We highlight three main observations. First, Transformers benefit most from CityMem: the city-agnostic Transformer baseline has the highest error among backbones (e.g., validation MSE $\approx 97.6$). Adding CityID slightly improves MAE but leaves MSE similar, whereas CityMem-Transformer reduces validation MSE to $\approx 64.7$ and MAE from $\approx 5.7$ to $\approx 4.2$, a relative improvement of roughly 30--35\% in MSE and 25\% in MAE. Second, graph backbones also see consistent gains. For the GNN backbone, CityID and CityMem both outperform the city-agnostic version, reducing MSE by about 8--9\%. For STGCN (Table~\ref{tab:traffic_full} and the supplemental STGCN aggregation), CityMem-STGCN achieves normalized error around 0.58 versus $\approx 0.86$ for the base model, a reduction of about one third. Third, effects on lower-capacity sequence models are smaller: GRU and TCN baselines already perform competitively and CityID/CityMem bring modest or no improvements. This supports our hypothesis that city-conditioned memory is most useful when the backbone has enough capacity to exploit the extra context.

\begin{table}[t]
    \centering
    \caption{Full-data traffic forecasting on METR-LA + PEMS-BAY. Means~$\pm$~std over three seeds.}
    \label{tab:traffic_full}
    \begin{tabular}{llll}
        \toprule
        backbone & model & MSE & MAE \\
        \midrule
        gnn & gnn\_base & 135.599 $\pm$ 2.398 & 7.107 $\pm$ 0.176 \\
        gnn & gnn\_cityid & 123.187 $\pm$ 4.423 & 6.454 $\pm$ 0.145 \\
        gnn & gnn\_citymem & 123.867 $\pm$ 2.899 & 6.610 $\pm$ 0.086 \\
        tf & tf\_base & 97.577 $\pm$ 3.621 & 5.718 $\pm$ 0.121 \\
        tf & tf\_cityid & 96.462 $\pm$ 3.109 & 5.658 $\pm$ 0.025 \\
        tf & tf\_citymem & 62.883 $\pm$ 5.159 & 4.231 $\pm$ 0.169 \\
        stgcn & base & 0.855 $\pm$ 0.229 & -- \\
        stgcn & cityid & 0.862 $\pm$ 0.234 & -- \\
        stgcn & citymem & 0.581 $\pm$ 0.006 & -- \\
        \bottomrule
    \end{tabular}
\end{table}

\subsection{Low-data traffic regimes}
\begin{figure}[htbp]
    \centering
    \includegraphics[width=0.85\linewidth,height=0.4\textheight,keepaspectratio]{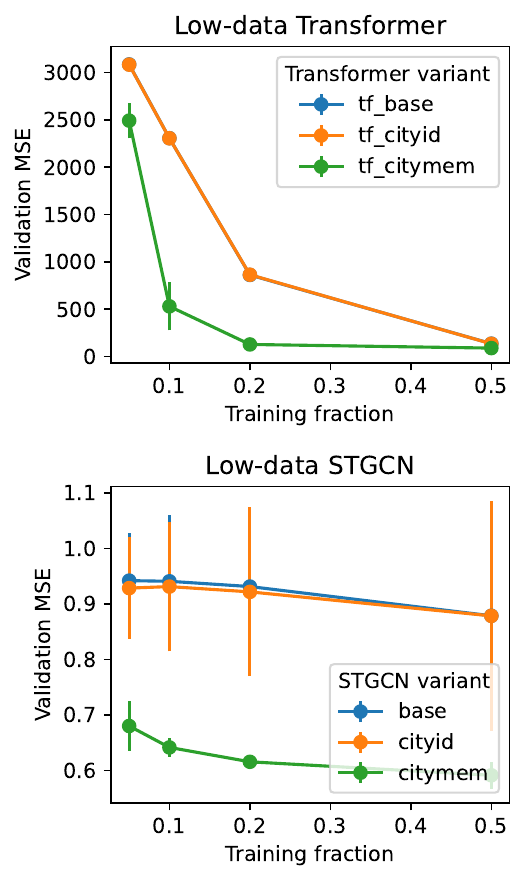}
    \caption{Low-data traffic forecasting performance. (a) Transformer variants as the available training fraction decreases. (b) STGCN variants under the same low-data settings.}
    \label{fig:traffic}
\end{figure}

Table~\ref{tab:traffic_low} and Figure~\ref{fig:traffic} show how error changes as we reduce the training data per city. Under severe data scarcity (5--10\%), all models degrade substantially, but city-conditioned versions degrade less. For example, at 10\% data, CityMem-Transformer's validation MSE is around 530--540, compared to over 2300 for the base Transformer, yielding a large relative improvement. CityMem-STGCN similarly reduces error by over 30\% across fractions. Under moderate scarcity (20--50\%), GRU and TCN gradually recover as more data is added, but still trail their full-data performance. CityMem-Transformer continues to outperform its baselines at all fractions, with especially strong gains at 20\% (roughly 4--5$\times$ lower MSE than base). For GNN and STGCN, CityID and CityMem consistently improve low-data performance, although the margin shrinks as data increases. Standard deviations across seeds remain small for most CityMem variants, suggesting that conditioning does not introduce optimization instability.

\begin{table}[t]
    \centering
    \caption{Low-data traffic forecasting on METR-LA + PEMS-BAY. Columns show the fraction of training data used. Means~$\pm$~std over three seeds.}
    \label{tab:traffic_low}
    \begin{tabular}{llll}
        \toprule
        backbone & model & 0.10 & 0.20 \\
        \midrule
        gnn & gnn\_base & 133.0 $\pm$ 8.5 & 138.3 $\pm$ 8.9 \\
        gnn & gnn\_cityid & 131.3 $\pm$ 9.4 & 129.1 $\pm$ 11.7 \\
        gnn & gnn\_citymem & 120.9 $\pm$ 7.1 & 124.9 $\pm$ 8.9 \\
        tf & tf\_base & 2305.7 $\pm$ 13.5 & 861.5 $\pm$ 43.7 \\
        tf & tf\_cityid & 2304.8 $\pm$ 17.3 & 864.1 $\pm$ 38.1 \\
        tf & tf\_citymem & 530.1 $\pm$ 254.9 & 127.7 $\pm$ 14.9 \\
        stgcn & base & 0.9 $\pm$ 0.1 & 0.9 $\pm$ 0.1 \\
        stgcn & cityid & 0.9 $\pm$ 0.1 & 0.9 $\pm$ 0.2 \\
        stgcn & citymem & 0.6 $\pm$ 0.0 & 0.6 $\pm$ 0.0 \\
        \bottomrule
    \end{tabular}
\end{table}

\subsection{Cross-city few-shot transfer}
\begin{figure}[htbp]
    \centering
    \includegraphics[width=0.85\linewidth,height=0.4\textheight,keepaspectratio]{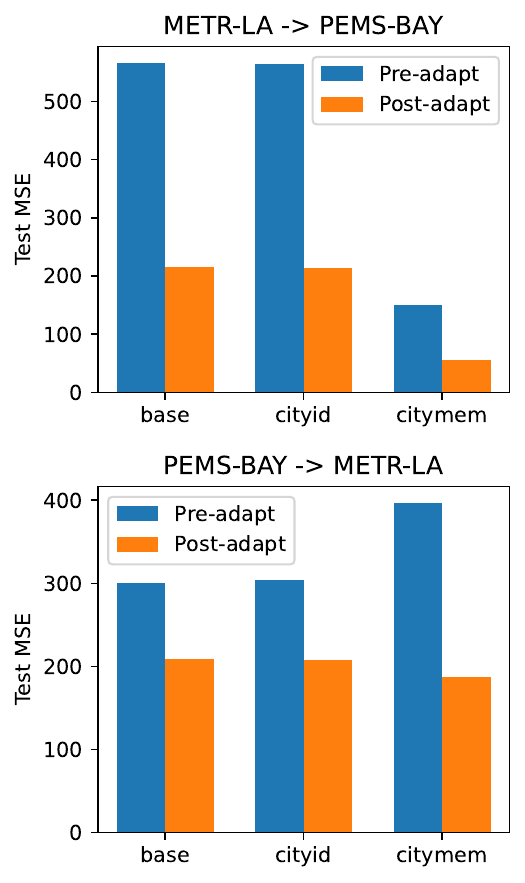}
    \caption{Cross-city few-shot transfer for Transformer backbones. Adaptation curves illustrate that CityMem improves both convergence speed and final error in both transfer directions.}
    \label{fig:transfer}
\end{figure}

We focus on Transformer-based models for cross-city transfer, as they benefit the most from CityMem. Table~\ref{tab:tf_xcity} reports pre- and post-adaptation test MSE/MAE for METR-LA $\rightarrow$ PEMS-BAY and PEMS-BAY $\rightarrow$ METR-LA. When trained on METR-LA and evaluated on PEMS-BAY without fine-tuning, CityMem-Transformer's error is already substantially lower than the city-agnostic and CityID baselines. After a short fine-tuning phase on the target city, all models improve, but CityMem-Transformer achieves the largest relative gains. Figure~\ref{fig:transfer} plots adaptation trajectories over gradient steps: CityMem-Transformer not only converges to a lower final error but also learns faster in the initial phase.

\begin{table}[t]
    \centering
    \caption{Cross-city few-shot transfer for Transformer backbones. ``Pre'' and ``Post'' denote error before and after adaptation on the target city. Means~$\pm$~std over three seeds.}
    \label{tab:tf_xcity}
    \begin{tabular}{lllll}
        \toprule
        train\_city & test\_city & model & Pre MSE & Post MSE \\
        \midrule
        METR-LA & PEMS-BAY & base & 566.0 $\pm$ 2.3 & 214.7 $\pm$ 1.3 \\
        METR-LA & PEMS-BAY & cityid & 564.3 $\pm$ 2.2 & 213.8 $\pm$ 0.7 \\
        METR-LA & PEMS-BAY & citymem & 150.3 $\pm$ 3.4 & 55.0 $\pm$ 0.3 \\
        PEMS-BAY & METR-LA & base & 299.9 $\pm$ 11.7 & 209.2 $\pm$ 1.1 \\
        PEMS-BAY & METR-LA & cityid & 303.7 $\pm$ 1.7 & 207.9 $\pm$ 3.9 \\
        PEMS-BAY & METR-LA & citymem & 396.8 $\pm$ 15.0 & 186.9 $\pm$ 1.5 \\
        \bottomrule
    \end{tabular}
\end{table}

\subsection{SIND multi-agent trajectories}
On SIND, we compare LSTM-Base and LSTM-CityID across full and low-data regimes. On the single-intersection dataset, CityID and Base achieve similar validation MSE and ADE/FDE, with differences under 2\%. CityID does not harm performance. When restricting training data to 5--50\%, CityID consistently yields slightly lower validation MSE and ADE/FDE than Base, especially at 10--20\% fractions (e.g., $\sim$5--8\% relative reductions). This indicates that even simple city embeddings can help sequence models disambiguate heterogeneous trajectories in low-data settings. We did not deploy CityMem on SIND in this work, both to keep the experimental matrix manageable and because pedestrian motion lacks an explicit graph structure; we leave memory designs tailored to such settings as future work.
\section{Analysis and Discussion}
\subsection{When does city conditioning help?}
Combining results across datasets, backbones, and regimes, a consistent picture emerges. For high-capacity sequence models (Transformer) and graph-based models such as GNNs and STGCNs, city conditioning---especially CityMem---substantially improves both full-data and low-data performance. It also yields strong benefits in cross-city transfer. For medium-capacity RNN/TCN backbones, the benefits are smaller and sometimes negligible, particularly in full-data regimes where the base model already fits well. For pedestrian trajectories, city-ID features help in low-data regimes but do not yet show the same magnitude of gains as in traffic, possibly due to task differences and the absence of graph structure. This suggests that city-conditioned memory is most valuable when the backbone has enough capacity to utilize additional context and when data per city is limited or heterogeneous.

\subsection{CityID vs CityMem vs Base}
Our experiments also shed light on the relative roles of CityID and CityMem. CityID is a strong baseline: for GNN and STGCN, CityID alone often provides most of the gains over Base by allowing the model to shift biases and scales per city. On SIND, CityID offers consistent small improvements at low data fractions. CityMem provides structured sharing: for Transformer and STGCN, CityMem significantly outperforms both Base and CityID, particularly in low-data and cross-city regimes. This indicates that a shared memory bank allows the model to distill and reuse global patterns that are not captured by city-specific embeddings alone. In a few configurations (e.g., GRU with very small data), CityMem does not help or is slightly worse than CityID, consistent with the intuition that adding memory increases model capacity and can overfit without enough data.

\subsection{Backbone capacity and data budget}
An interesting interaction emerges between backbone capacity and data budget. With large data, high-capacity models may already extract most of the useful structure from each city independently; city conditioning then offers moderate gains. With limited data, high-capacity models are more prone to overfitting, but CityMem acts as a form of cross-city regularization: memory slots aggregated from multiple cities encourage sharing of reusable patterns (e.g., morning/evening rush-hour profiles). For low-capacity models (e.g., two-layer GRUs), the bottleneck remains the backbone itself; adding city conditioning brings smaller marginal improvements. These findings complement prior work on graph-based and meta-learning approaches~\cite{li2018dcrnn,yu2018stgcn,wu2019graphwavenet,wu2020connecting,bai2020agrn,yao2019metast,panagopoulos2021stmetanet,jin2022selective,wu2021survey}, suggesting that even simple conditioning layers can meaningfully reshape the bias--variance trade-off in multi-city forecasting.

\subsection{Qualitative behavior of CityMem}
We perform a preliminary qualitative analysis of learned memory slots for CityMem-Transformer and CityMem-STGCN (visualizations in the appendix). We observe that certain slots are consistently activated during rush-hour periods across both METR-LA and PEMS-BAY, suggesting they capture global traffic patterns such as ``morning commuting into downtown''. Other slots show city-specific preferences---for example, slots that are more active in PEMS-BAY's evening rush hour than in METR-LA---indicating that the attention mechanism naturally balances global reuse and city-specific specialization. During cross-city adaptation, attention shifts gradually from source-dominated to target-dominated slots, providing an interpretable view of how the model reuses and reweights shared patterns. While these observations are preliminary, they support our interpretation of CityMem as a shared pool of reusable spatio-temporal motifs.

\section{Related Work}
\subsection{Spatio-temporal forecasting and graph neural networks}
Deep spatio-temporal models have become the de facto standard for traffic forecasting. DCRNN introduces diffusion convolution within an RNN to capture directed graph dynamics~\cite{li2018dcrnn}. STGCN uses localized graph convolutions combined with temporal convolutions~\cite{yu2018stgcn}. Graph WaveNet augments graph learning with dilated temporal convolutions~\cite{wu2019graphwavenet}, and MTGNN further unifies multivariate time-series forecasting with learned graphs~\cite{wu2020connecting}. Adaptive graph RNNs extend these ideas by learning dynamic adjacency matrices~\cite{bai2020agrn}. Surveys on spatial-temporal GNNs provide broader context for these architectures~\cite{wu2021survey}. Our work is orthogonal to these backbones: CityCond neither changes their core message-passing mechanisms nor their loss functions, but sits on top as a city-conditioned adapter.

\subsection{Cross-city transfer and meta-learning}
RegionTrans (Wang et al.) proposed regional transfer learning for deep spatio-temporal prediction by aligning similar regions across cities~\cite{yao2019regiontrans}. MetaST introduced a global pattern-based spatio-temporal memory and a meta-learning framework for transferring to data-scarce target cities~\cite{yao2019metast}. ST-MetaNet (Pan et al.) used deep meta learning for urban traffic prediction~\cite{panagopoulos2021stmetanet}. Selective Cross-City Transfer (CrossTReS) further studied which cities to transfer from and how much to transfer via region re-weighting~\cite{jin2022selective}. These methods highlight the importance of cross-city sharing but typically require complex training procedures (e.g., inner-loop optimization, region alignment). By contrast, CityCond is a single layer that can be attached to existing backbones and trained with standard supervised learning, while still capturing useful cross-city structure via shared memory.

\subsection{Trajectory forecasting at signalized intersections and SIND}
Pedestrian trajectory prediction has seen rapid progress with RNNs and generative models. Social GAN uses a GAN framework to generate socially plausible trajectories~\cite{gupta2018socialgan}. Trajectron++ models multi-agent interactions and dynamic feasibility with conditional variational auto-encoders~\cite{salzmann2020trajectronpp}. Social-STGCNN applies spatio-temporal GCNs to human trajectories~\cite{mohamed2020socialstgcnn}. The SIND drone dataset provides multi-agent trajectories from a signalized intersection in Tianjin~\cite{wei2024sind}. We use SIND primarily as a complementary mobility task to test the generality of CityID-style conditioning in a non-graph setting. Designing memory mechanisms that explicitly account for human--human interactions remains an interesting direction.

\section{Conclusion and Future Directions}
We introduced CityCond, a light-weight city-conditioned memory layer that can be attached to a variety of spatio-temporal backbones. Through a unified experimental study on METR-LA, PEMS-BAY, and SIND, we showed that CityCond---especially its CityMem variant---consistently improves multi-city traffic forecasting in full-data, low-data, and cross-city few-shot regimes, with the largest gains for high-capacity backbones such as Transformers and STGCNs. Simple city-ID embeddings also help in pedestrian trajectory forecasting, particularly when data per scene is limited. Rather than proposing a monolithic new architecture, we advocate viewing city-conditioned memory as a reusable design pattern for scaling forecasting models across many urban systems under realistic resource constraints. In practice, CityCond can be implemented as a small module that wraps existing models, making it easy to integrate into production systems. Future work includes: extending CityCond to more cities and modalities (e.g., ride-hailing demand, bike sharing); incorporating richer city descriptors (POIs, demographics, policies) into the city encoder; exploring finer-grained memories at the region or sensor level; and deploying CityCond in online or continual-learning settings where cities evolve over time.

\FloatBarrier

\end{document}